%% file: lrec-coling2024-example.tex
\documentclass[10pt, a4paper]{article}

\usepackage{pifont}
\usepackage{enumitem}
\usepackage{arydshln}
\usepackage{booktabs}
\usepackage{multirow}
\usepackage[noend]{algpseudocode}
\usepackage{algorithm}
\usepackage[dvipsnames]{xcolor}
\usepackage{tcolorbox}
\usepackage{tikz}
\usepackage{pgfplots}
\usepackage[]{lrec-coling2024} % this is the new style

\title{RankPrompt: Step-by-Step Comparisons Make \\ Language Models Better Reasoners}

\name{
Chi Hu$^{1,2*\thanks{*Work done while the author was interning at Alibaba.}}$, 
Yuan Ge$^1$, Xiangnan Ma$^1$, Hang Cao$^1$, \\ 
\textbf{Qiang Li$^2$,} \textbf{Yonghua Yang$^2$,} \textbf{Tong Xiao$^{1,3}$}, \textbf{Jingbo Zhu$^{1,3\dag}$\thanks{$^\dag$Corresponding author.}}
}

\address{
$^1$School of Computer Science and Engineering,\\ Northeastern University, Shenyang, China\\
$^2$Alibaba Group \\
$^3$NiuTrans Research, Shenyang, China \\
         \{huchinlp, geyuanqaq\}@gmail.com \\
         \{xiaotong, zhujingbo\}@mail.neu.edu.cn\\
         \{lq178896, huazai.yyh\}@taobao.com
}

\abstract{
Large Language Models (LLMs) have achieved impressive performance across various reasoning tasks. However, even state-of-the-art LLMs such as ChatGPT are prone to logical errors during their reasoning processes. Existing solutions, such as deploying task-specific verifiers or voting over multiple reasoning paths, either require extensive human annotations or fail in scenarios with inconsistent responses. To address these challenges, we introduce RankPrompt, a new prompting method that enables LLMs to self-rank their responses without additional resources. RankPrompt breaks down the ranking problem into a series of comparisons among diverse responses, leveraging the inherent capabilities of LLMs to generate chains of comparison as contextual exemplars. Our experiments across 11 arithmetic and commonsense reasoning tasks show that RankPrompt significantly enhances the reasoning performance of ChatGPT and GPT-4, with improvements of up to 13\%. Moreover, RankPrompt excels in LLM-based automatic evaluations for open-ended tasks, aligning with human judgments 74\% of the time in the AlpacaEval dataset. It also exhibits robustness to variations in response order and consistency. Collectively, our results validate RankPrompt as an effective method for eliciting high-quality feedback from language models.
\\ \newline \Keywords{Language Modeling, Reasoning, Model Feedback} }

\pgfplotsset{compat=1.18}
\begin{document}

\maketitleabstract

\section{Introduction}
Reasoning ability is a fundamental aspect of human intelligence, crucial for tasks such as mathematical problem-solving \cite{koncel-kedziorski-etal-2016-mawps, ling-etal-2017-program} and questions-answering \cite{talmor-etal-2019-commonsenseqa, geva-etal-2021-aristotle}. Recent advancements show that Large Language Models (LLMs) \cite{Brown2020LanguageMA, Thoppilan2022LaMDALM, Chowdhery2022PaLMSL, Ouyang2022TrainingLM} can demonstrate remarkable reasoning abilities when guided by Chain-of-Thought (CoT) prompting \cite{Wei2022ChainOT, Kojima2022LargeLM}. This technique provides LLMs with prompts, such as ``\textit{Let's think step by step}”, to facilitate the generation of a sequence of intermediate steps before arriving at the final result. CoT prompting has yielded impressive performance across a variety of tasks, including arithmetic, commonsense, and symbolic reasoning \cite{Wei2022EmergentAO, Zhang2022AutomaticCO, Suzgun2022ChallengingBT, Zhou2022LeasttoMostPE}.

\input{tables/intro}

Despite their success, LLMs often make logical mistakes during the reasoning process \cite{Kojima2022LargeLM, Turpin2023LanguageMD, Lightman2023LetsVS}. As shown in Table \ref{tab:intro}, when solving algebra problems, a language model may provide wrong inferences or omit pivotal steps, leading to incorrect final results. One potential solution is to use task-specific verifiers to validate each step \cite{Cobbe2021TrainingVT, li-etal-2023-making, Lightman2023LetsVS}. However, it requires substantial labeled data for training, which is costly and time-consuming. An alternative is to sample a variety of reasoning paths and aggregate the results via majority voting \cite{Wang2022SelfConsistencyIC, Fu2022ComplexityBasedPF}. This method can alleviate the impact of individual errors and lead to more accurate predictions \cite{ Huang2022TowardsRI, Huang2022LargeLM}. Nevertheless, this aggregate voting strategy ignores intermediate steps, lacks interpretability, and struggles with inconsistent answers, as illustrated in Table \ref{tab:intro}. Therefore, it is crucial to develop a robust, interpretable technique that can effectively distinguish among multiple reasoning paths, thereby augmenting the reasoning capabilities of LLMs. 

\begin{figure*}
	\centering
	\includegraphics[width=160mm]{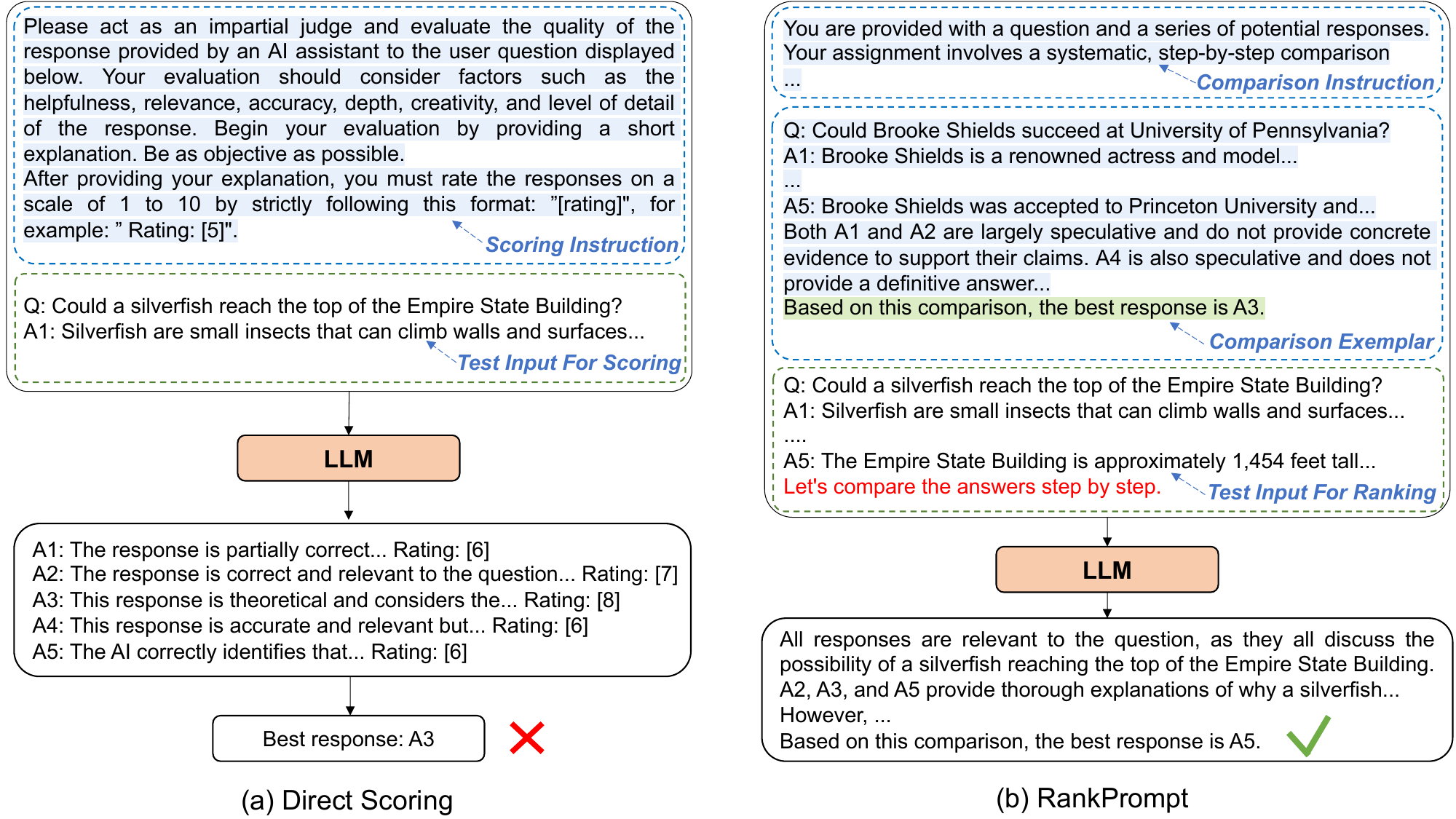}
	\caption{An overview of Direct Scoring \cite{Zheng2023JudgingLW} (left) and RankPrompt (right). Direct Scoring independently assigns scores to each candidate, whereas RankPrompt ranks candidates through a systematic, step-by-step comparative evaluation. We present the detailed instructions for comparison in Table \ref{tab:ranking_template} and describe the construction of comparison exemplars in Section \ref{sec:exemplars}.}
 \label{fig:main_method}
\end{figure*}

In response to these challenges, we introduce RankPrompt, a novel prompting method for LLM-based reasoning. Unlike previous methods, RankPrompt generates diverse reasoning paths and instructs LLMs to select the optimal one. As illustrated in Figure \ref{fig:main_method}, RankPrompt diverges from the well-established Direct Scoring method \cite{Zheng2023JudgingLW}, which assesses candidates individually. Instead, our approach directs LLMs to perform a comparative evaluation of candidates through two essential components: \textit{step-aware comparison instructions} and \textit{comparison exemplars}. The former decomposes the ranking problem into a series of comparisons, using instructions such as ``Let's compare the answers step by step”. The latter component, comparison exemplars, leverages the few-shot learning capabilities of LLMs to improve ranking performance further. In contrast to previous methods requiring manual design of exemplars \cite{Wei2022ChainOT, Wang2022SelfConsistencyIC}, our approach tasks LLMs with generating multiple chains of comparisons and selecting the chains yielding correct ranking results as exemplars. These exemplars guide LLMs to systematically compare different paths, thereby reducing the requirement for labeled data and minimizing human intervention.

We evaluate RankPrompt across various arithmetic, commonsense, and symbolic reasoning benchmarks using ChatGPT. Empirical results demonstrate that RankPrompt consistently outperforms CoT prompting, achieving an improvement of up to 13\% on the AQUA-RAT \cite{ling-etal-2017-program} data. On more challenging tasks from BIG-Bench-Hard \cite{Suzgun2022ChallengingBT}, RankPrompt boosts the performance of GPT-4 \cite{gpt-4-technical-report}, with gains ranging from 5.2\% to 9.2\%. While our primary focus is on reasoning tasks, RankPrompt also excels in assessing open-ended generation. Specifically, it sets a new standard for LLM-based automatic evaluation by achieving a 74\% agreement rate with human judgment on the AlpacaEval set. Remarkably, these impressive results can be obtained using a single exemplar, which underscores the efficacy of RankPrompt. Our analysis demonstrates that RankPrompt is robust to the order of candidate answers. Overall, our findings highlight the importance of considering intermediate steps in ranking tasks and establish RankPrompt as a promising approach for improving LLM-based reasoning.

\section{Related Work}

There is a surge in research interest in the field of LLMs due to their exceptional performance across a wide array of tasks \cite{Brown2020LanguageMA, Thoppilan2022LaMDALM, Chowdhery2022PaLMSL, Hoffmann2022TrainingCL, OpenAI2023GPT4TR}. A key aspect of LLMs is their emergent abilities when provided with appropriate context \cite{Wei2022EmergentAO, OpenAI2023GPT4TR, Zhao2023ASO}, leading to their potential use in reasoning and automatic evaluation. Here, we briefly discuss related work in the two fields.

\paragraph{LLMs as Reasoners.} 
Reasoning with Language Models (LLMs) has become a popular research topic. One promising methodology is Chain-of-Thought (CoT) prompting, which encourages LLMs to generate a chain of reasoning steps (called a reasoning path) before delivering a final answer. This approach has been shown to improve the performance of LLMs across various tasks. CoT prompting optimization generally falls into two categories. The first focuses on enhancing the quality of individual reasoning paths through prompt engineering. For example, \citet{Kojima2022LargeLM} find that specific trigger words can significantly improve the zero-shot reasoning performance of LLMs. Meanwhile, \citet{Fu2022ComplexityBasedPF} demonstrate that incorporating complex exemplars into prompts can notably enhance the few-shot reasoning capabilities of LLMs. However, these methods often necessitate careful design and manipulation of prompts. The second category involves generating multiple reasoning paths and applying specific strategies to select the most effective one. For example, \citet{Wang2022SelfConsistencyIC} use majority voting to select the final results, while \citet{li-etal-2023-making} and \citet{Lightman2023LetsVS} train step-aware verifiers to validate reasoning steps. Nonetheless, these methods also face challenges. Majority voting lacks interpretability and is prone to inconsistent final answers, while training verifiers requires a significant amount of labeled data. Our method addresses these limitations while complementing existing strategies for improving the quality of individual reasoning paths.

\paragraph{LLMs as Evaluators.} Recent studies have explored the potential of LLMs in evaluating and refining their outputs. For instance, \citet{Liu2023GEvalNE} and \citet{Wang2023IsGoodCA} utilize LLMs to assess the quality of text generation tasks such as summarization and machine translation. Similarly, \citet{Madaan2023SelfRefineIR} use LLMs to iteratively refine outputs for more complex tasks, such as acronym generation and code optimization. \citet{Dubois2023AlpacaFarmAS} and \citet{Zheng2023JudgingLW} show that, when equipped with carefully designed prompts, GPT-4 exhibits a high correlation with human preferences in judging the quality of open-ended text generation. It is established that LLM-based evaluators are cost-effective and efficient alternatives to crowd annotators \cite{fu2023gptscore, Liu2023GEvalNE, Dubois2023AlpacaFarmAS, Zheng2023JudgingLW}. However, the challenge lies in designing effective prompts to elicit the ranking ability of LLMs, often requiring significant human effort and extensive interactions with LLMs \cite{Liu2023GEvalNE, Wang2023Fair, Wang2023IsGoodCA}. In this paper, we extend this line of research by developing a method that leverages LLMs to automatically generate exemplars for ranking, significantly reducing the need for human intervention. Our study also contributes to understanding how LLMs can be effectively utilized for reasoning and automatic evaluation tasks.

\section{Method}
This section introduces RankPrompt, a two-stage prompting framework for reasoning tasks. In the first stage, we generate multiple diverse reasoning paths, each potentially leading to a unique outcome. Our focus primarily lies in the second stage, where we re-rank these reasoning paths by comparing their steps and selecting the optimal one as the final answer.

\input{tables/ranking_template.tex}

\subsection{Candidate Generation}
\label{sec:candidate_generation}
The generation and aggregation of multiple reasoning paths have been proven to boost the performance of reasoning models \cite{Wang2022SelfConsistencyIC, Fu2022ComplexityBasedPF}. This process is similar to ensemble learning, a well-established machine learning method that combines the outputs of multiple models to improve overall accuracy and robustness against individual errors \cite{10.5555/648054.743935}.

Given a question $q$, we generate $n$ reasoning paths $\mathbf{p} = ({p_1, p_2, \ldots, p_n})$, each potentially leading to a different final answer. We use few-shot CoT prompting  \cite{Wei2022ChainOT, Wang2022SelfConsistencyIC} to generate these reasoning paths and apply temperature sampling \cite{Ficler2017ControllingLS, Fan2018HierarchicalNS} to encourage diversity among the generated paths. Each reasoning path $p_i$ (where $i \in {1, \ldots, n}$) corresponds to a set of final answers $\mathbf{r} = ({r_1, r_2, \ldots, r_n})$. We refer to the pairs $(p_i, r_i)$, where each reasoning path $p_i$ corresponds to a final answer $r_i$, as the \textit{candidates} for question $q$. Hence, the candidate generation process results in a set of candidates $C_q = \{(p_1, r_1), (p_2, r_2), \ldots, (p_n, r_n)\}$ for each question $q$. We then use the candidate set $C_q$ as the input for the subsequent ranking process.

\subsection{Candidate Ranking}
\label{sec:candidate_ranking}
\subsubsection{Comparative Evaluation of Reasoning Steps}
\label{sec:step_aware_comparison_instructions}
 A common approach to candidate ranking is to evaluate each candidate individually \cite{Zheng2023JudgingLW, Wang2023Fair}, a strategy we refer to as \textbf{Direct Scoring} (Figure \ref{fig:main_method}(a)). However, such an approach often fails to account for the relative quality of different reasoning paths. For instance, LLMs such as ChatGPT often assign identical scores to candidates with similar reasoning steps, regardless of their differing outcomes \cite{Dubois2023AlpacaFarmAS, Zheng2023JudgingLW}. 

To address this limitation, we introduce a comparative evaluation method, which concatenates all candidate reasoning paths with the original question to form the ranking input. This input is then processed by a ranking model, such as ChatGPT, guided by a step-aware comparison instruction. As presented in Table \ref{tab:ranking_template}, the comparison instruction directs the model to execute a sequential comparison process before giving the conclusion. It also clarifies the required output format.

\begin{algorithm}[th]\small
\caption{Creation of Comparison Exemplars}\label{alg:comparison_exemplars}
\begin{algorithmic}[1]
\Require Labeled data set $D=\{(q_1,a_1), \ldots, (q_k,a_k)\}$, where $q_i$ is a question and $a_i$ is the correct answer, empty exemplar set $E$
\Ensure Comparison exemplar set $E = (e_1, \ldots, e_k)$
\Procedure{CreateExemplars}{$D$}
\For{each data point $(q_j,a_j)$ in $D$} 
\State Generate a diverse candidate set $C_{q_j}$ for $q_j$
\State Initialize $e_j$ as an empty exemplar
\While{$e_j$ has not been created for $q_j$}
\State Generate a comparison chain $c_j$ using Zero Ranking with $(q_j, C_{q_j})$
\If{$c_j$ meets selection criteria} 
\State Append $e_j = (q_j, C_{q_j}, c_j)$ to $E$
\State \textbf{break}
\EndIf
\EndWhile
\EndFor
\State \textbf{return} $E$
\EndProcedure%
\end{algorithmic}
\end{algorithm}

However, relying solely on comparison instructions, which we refer to as \textbf{Zero Ranking}, does not fully leverage the in-context learning capabilities of LLMs \cite{Brown2020LanguageMA, Wei2022ChainOT}. The Zero Ranking method can sometimes lead to irrelevant outputs, failure to adhere to the desired output format, or only a partial consideration of candidates \cite{Sun2023IsCG, Qin2023LargeLM}. To address these issues, we enhance the ranking capabilities of LLMs by incorporating comparison exemplars, as shown in Figure \ref{fig:main_method}(b).

\subsubsection{Construction of Comparison Exemplars}
\label{sec:exemplars}
To fully exploit the in-context learning capabilities of Language Model Machines (LLMs), we enhance the instructions with high-quality examples. However, creating such examples can be a challenging and time-consuming task \cite{Lu2021FantasticallyOP, Liu2021WhatMG, Fu2022ComplexityBasedPF}. To address this issue, we propose an automatic method for generating comparison examples, as shown in Algorithm \ref{alg:comparison_exemplars}.
\input{tables/main_results}

Algorithm \ref{alg:comparison_exemplars} initiates by iterating through a labeled dataset $D$, creating a candidate set $C_{q_j}$ for every question $q_j$. It then continuously produces comparison chains using Zero Ranking until it identifies a chain that meets the selection criteria. Echoing the approach of \citet{ZelikmanWMG22}, we select the comparison chain that accurately leads to the answer $a_j$. This chosen chain, along with the question and its candidate set, forms an exemplar $e_j$, which is subsequently added to the exemplar collection $E$. This procedure is repeated for each question until a suitable chain is found. Compared to previous methods, our approach requires only a minimal amount of labeled data for each task. In Section \ref{sec:analysis}, we delve into the effects of exemplar selection on the efficacy of the ranking process.

\section{Experiment}
\subsection{Experimental Setups}
\paragraph{Models.}
We evaluate our method using state-of-the-art LLMs, including \texttt{gpt-3.5-turbo} and \texttt{gpt-4}, via the OpenAI API\footnote{\url{https://platform.openai.com/docs/api-reference}}. Additionally, we test a variant of ChatGPT, \texttt{gpt-3.5-turbo-16k}, which supports an input length of up to 16K, to analyze the impact of varying numbers of exemplars and candidates. Our experimental evaluations were carried out between August 1, 2023, and October 1, 2023.

\paragraph{Tasks and Datasets.}
We conduct experiments with \texttt{gpt-3.5-turbo} across 8 widely-used reasoning tasks, spanning arithmetic, commonsense, and symbolic reasoning. For arithmetic reasoning, we use 4 math word problem datasets: AQUA-RAT \cite{ling-etal-2017-program}, ASDiv \cite{Miao2020ADC}, GSM8K \cite{Cobbe2021TrainingVT}, and SVAMP \cite{Patel2021AreNM}. For commonsense reasoning, which requires multi-step problem-solving, we utilize ARC Challenge \cite{Clark2018ThinkYH}, CommonsenseQA \cite{talmor-etal-2019-commonsenseqa}, and StrategyQA \cite{geva-etal-2021-aristotle}. We evaluate symbolic reasoning with the Last Letter Concatenation task \cite{Wei2022ChainOT}. Given the high API cost\footnote{\url{https://openai.com/pricing}}, we reserve \texttt{gpt-4} for 3 challenging reasoning tasks from BIG-Bench-Hard \cite{Suzgun2022ChallengingBT}: Causal Judge, Logical Deduction Seven Objects, and Formal Fallacies. Following \citet{Wang2022SelfConsistencyIC}, we report the accuracy on the test set for all tasks except CommonsenseQA, where we use the validation set. Additionally, we test RankPrompt on AlpacaEval \cite{Dubois2023AlpacaFarmAS}, a benchmark for measuring LLM-based automatic evaluation of open-ended generation. The benchmark comprises 805 instructions, each with a pair of responses and 4 human preferences. We compare different methods using \texttt{gpt-4} and report the level of agreement with human preferences.

\paragraph{Candidate Generation Setups.}
For a fair comparison, we employ the same prompts created by \citet{Wei2022ChainOT} and \citet{Suzgun2022ChallengingBT} for candidate generation. We use a temperature of 0.7 to generate 5 reasoning paths as candidates. We restrict our selection to 5 candidates, as increasing this number yields only marginal performance improvements. Additionally, adding more candidates would increase the API costs due to context expansion. In Section \ref{exp:num_candidates}, we thoroughly analyze the impact of candidate numbers on the results.

\paragraph{Ranking Setups.}
We leverage language models to rank their outputs. For each task, a task-specific comparison exemplar is generated using the same model utilized for candidate generation. These exemplars systematically evaluate 5 unique candidate responses, ultimately guiding models to the correct answer. Following this, we integrate these exemplars into the ranking template, as detailed in Table \ref{tab:ranking_template}. Despite the diverse nature of tasks, we maintain a uniform application of comparison instructions and task-specific exemplars, introducing minor modifications to the output format depending on the task type. We restrict our use of comparison exemplars to a single one, as our findings suggest that increasing the number of exemplars has a negligible effect on improving performance but significantly extends the input, often exceeding the maximum length limit of OpenAI models. In Section \ref{sec:analysis}, we conduct a comprehensive examination of how various facets of comparison exemplars influence the final performance.

\paragraph{Baselines.}
We compare our methods with 4 baseline methods: CoT Prompting \cite{Wei2022ChainOT}, Majority Voting \cite{Wang2022SelfConsistencyIC}, Direct Scoring \cite{Zheng2023JudgingLW}, and Zero Ranking. Majority Voting selects the answer that appears most frequently. At the same time, Direct Scoring uses the prompt template proposed by \citet{Zheng2023JudgingLW} to evaluate candidates independently, soliciting Large Language Models (LLMs) to rank candidates on a scale from 1 to 10. Zero Ranking, the final baseline, employs the comparison instruction shown in Table \ref{tab:ranking_template}, but excludes the comparison exemplars.

\subsection{Main Results}
\label{sec:main_results}
Table \ref{table:main_results} summarizes the experimental results on 8 reasoning tasks using \texttt{gpt-3.5-turbo}. The CoT Prompting method stands out as it employs greedy decoding at a temperature of 0, while other methods sample 5 candidates at a temperature of 0.7. We also report the \textit{oracle} results, which represent the upper bounds of re-ranking, identified by selecting the optimal response from all possible candidates.

The results demonstrate that both the voting and ranking methods considerably outperform CoT Prompting. Majority Voting and Direct Scoring show similar performance (averaging 78.91 and 78.77, respectively), slightly falling behind Zero Ranking (which averages 79.44). Notably, RankPrompt emerges as the best-performing method, achieving the highest scores in all categories except for ARC, where all methods demonstrate comparable performance. We also find that RankPrompt is more effective for challenging tasks such as AQuA-RAT, GSM8K, and CSQA. In particular, it significantly surpasses other methods on the AQuA-RAT dataset, achieving a 13\% improvement over CoT Prompting. These findings highlight the importance of incorporating comparison exemplars in the ranking process. Additionally, the Oracle results signal considerable potential for future enhancements in ranking methods.

\input{tables/bbh_results}

\subsection{Results on More Challenging Tasks}
\label{sec:bbh_exp}
To further probe the performance on complex tasks, we test various methods on 3 challenging BIG-Bench Hard (BBH) tasks using \texttt{gpt-4}. We apply the prompt templates created by \citet{Suzgun2022ChallengingBT} for the CoT Prompting baseline and generate candidates with identical settings as described in Section 
\ref{sec:main_results}. 

\input{tables/alpaca_results}

Table \ref{tab:bbh_results} shows the experimental results. We observe that Majority Voting beats Direct Scoring, yet falls short when compared to Zero Ranking. RankPrompt emerges as superior over all other methods, achieving performance improvements ranging from 5.2\% to 9.2\% compared to CoT Prompting. These results validate that RankPrompt is highly effective for complex reasoning tasks.

\begin{figure}[t!]
    \centering
    \small
    \includegraphics[scale=0.25]{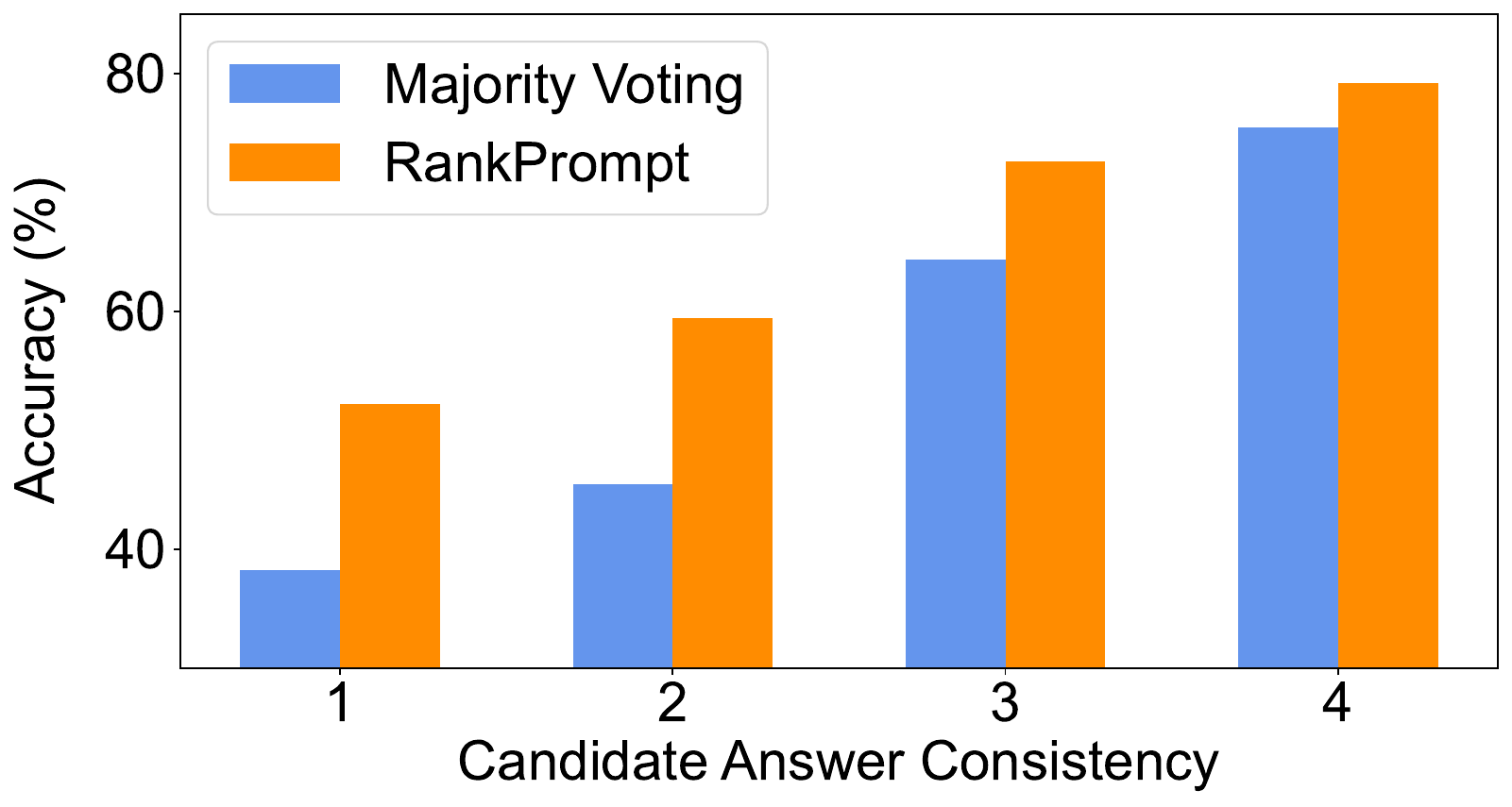}
    \caption{RankPrompt performs much better than majority voting when the candidate answers are inconsistent. The results are obtained on AQuA-RAT over 5 candidates using \texttt{gpt-3.5-turbo}.}
    \label{fig:consistency}
\end{figure}

\subsection{Results on Inconsistent Candidates}
\label{sec:inconsistency_exp}
The results mentioned above show that RankPrompt consistently outperforms Majority Voting across various tasks. We delve deeper into the results of AQUA-RAT by categorizing candidates based on their consistency. We determine consistency by the frequency of major answers among the candidates. Suppose we have $n$ candidates. When all candidates are identical, the consistency reaches $n$, eliminating the need for re-ranking. Conversely, in the most challenging scenario where all candidates are unique, the number of consistent answers drops to 1. We conduct experiments with \texttt{gpt-3.5-turbo} on the AQUA-RAT dataset, maintaining the same settings as in Section \ref{sec:main_results}.

Figure \ref{fig:consistency} illustrates that RankPrompt and Majority Voting exhibit high accuracy when the answer candidates are consistent, especially when there are more than 3 consistent answers. However, the performance dramatically drops when the number of consistent answers is less than 3. Despite this decrease, RankPrompt notably outperforms the voting method. These observations validate our motivation that relying solely on the final answer does not guarantee accurate identification of the optimal candidate.

\subsection{Results on Automatic Evaluation}
\label{sec:automatic_evaluation_exp}
In this section, we delve deeper into the effectiveness of RankPrompt by examining its performance in automatic evaluation tasks. We test RankPrompt on the AlpacaEval benchmark introduced by \citet{Dubois2023AlpacaFarmAS}. This benchmark comprises a test set of 805 instructions, each accompanied by pairs of responses, designed to assess the instruction-following abilities of language models. Our comparison incorporates Direct Scoring \cite{Zheng2023JudgingLW}, AlpacaFarm, AlpacaEval \cite{Dubois2023AlpacaFarmAS}, and Zero Ranking. We assess the performance of each method by calculating the agreement rate with the majority of human preferences, a critical metric for understanding how well each approach aligns with human judgment. Additionally, we present a detailed analysis of the costs associated with each method, including the expenses related to human annotations as reported by \citet{Dubois2023AlpacaFarmAS}. We experiment with \texttt{gpt-4} and present the results in Table \ref{tab:alpaca_results}. RankPrompt outperforms all other methods, achieving a 74.33\% agreement rate with human evaluators—Direct Scoring, however, trails by a significant 10\% margin. Interestingly, LLM-based evaluators not only yield superior results but also reduce cost by more than 90\% compared to crowd-sourced annotators. These findings underscore the critical role of appropriate instructions and exemplars when comparing candidate answers.

\section{Analysis}
\label{sec:analysis}
In this section, we thoroughly study the factors that influence ranking performance. Specifically, we examine the effect of exemplars and candidate reasoning paths on ranking outcomes. We also analyze the errors produced by different methods in the complex arithmetic reasoning task. Through this analysis, we aim to deepen the understanding of our proposed method.

\subsection{Impact of Comparison Exemplars}
\begin{figure}[t]
    \centering
    \includegraphics[scale=0.25]{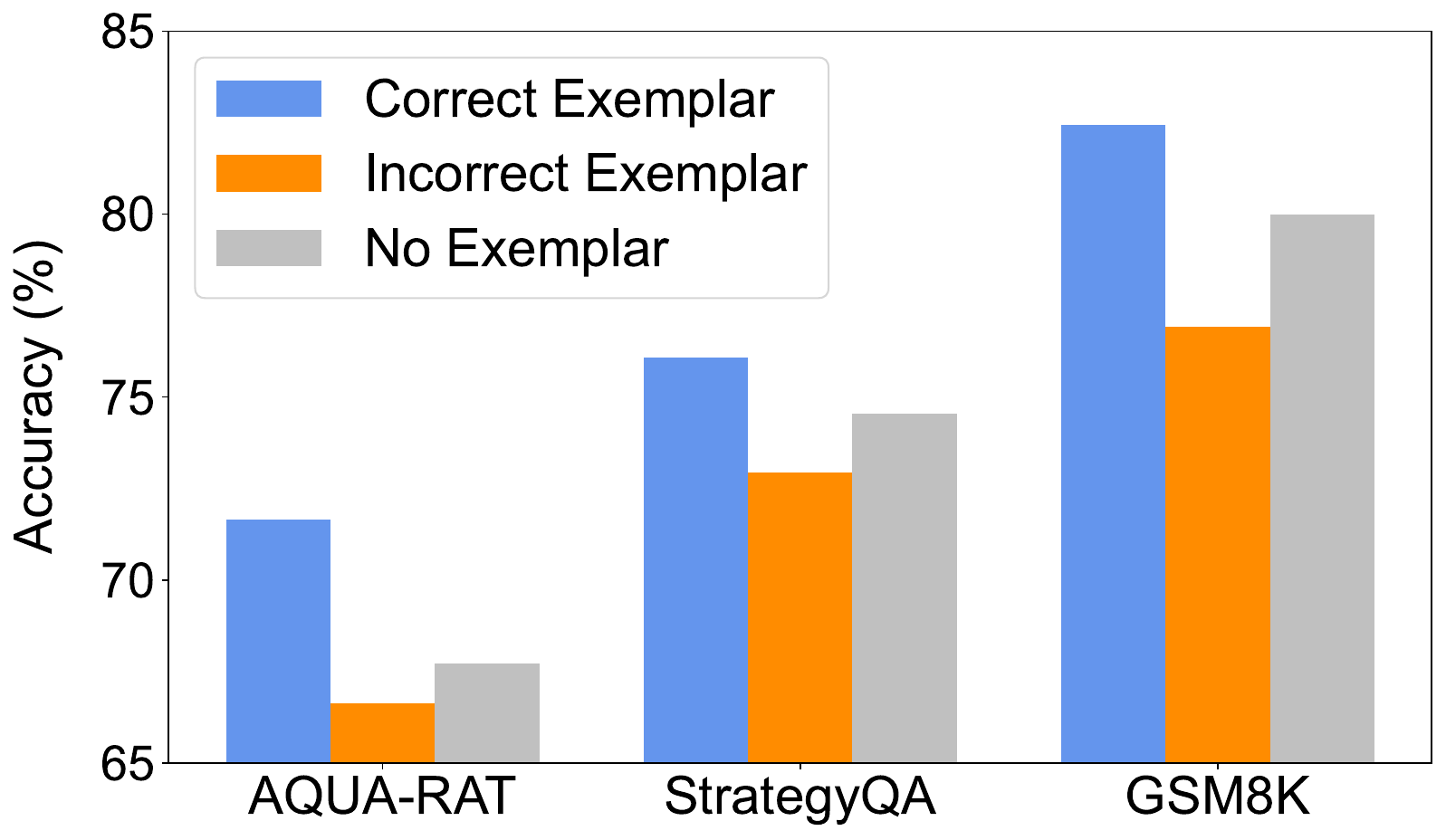}
    \caption{Performance of RankPrompt with a correct example vs. an incorrect example when ranking over 5 candidates. The results are obtained with \texttt{gpt-3.5-turbo}.}
    \label{fig:correctness}
\end{figure}

\paragraph{Exemplar correctness is the key to the performance of RankPrompt.}
A fundamental component of RankPrompt is its selection of comparison paths that yield the correct answers. It has been established that, in almost all cases, the intermediate steps generated by LLMs are also correct when the final result of inference is accurate \cite{wang-etal-2023-towards}. Here, we aim to shed light on how the accuracy of the comparison exemplars influences the overall effectiveness of our method. In the experiments, we condition \texttt{gpt-3.5-turbo} with no exemplars, correct exemplars, and incorrect exemplars, respectively. We adhere to the settings specified in Section \ref{sec:main_results} for candidate generation and apply a single exemplar for ranking. Our evaluation comprises 3 tasks: GSM8K, AQUA-RAT, and StrategyQA. As illustrated in Figure \ref{fig:correctness}, the use of incorrect exemplars invariably compromises the performance of the ranking, particularly in more challenging tasks such as AQUA-RAT. On the other hand, the application of correct exemplars consistently enhances the accuracy when contrasted with the use of no exemplars or inconsistent ones. These findings establish that choosing the correct exemplars is essential for RankPrompt.

\begin{figure}[t]
    \centering
    \includegraphics[scale=0.25]{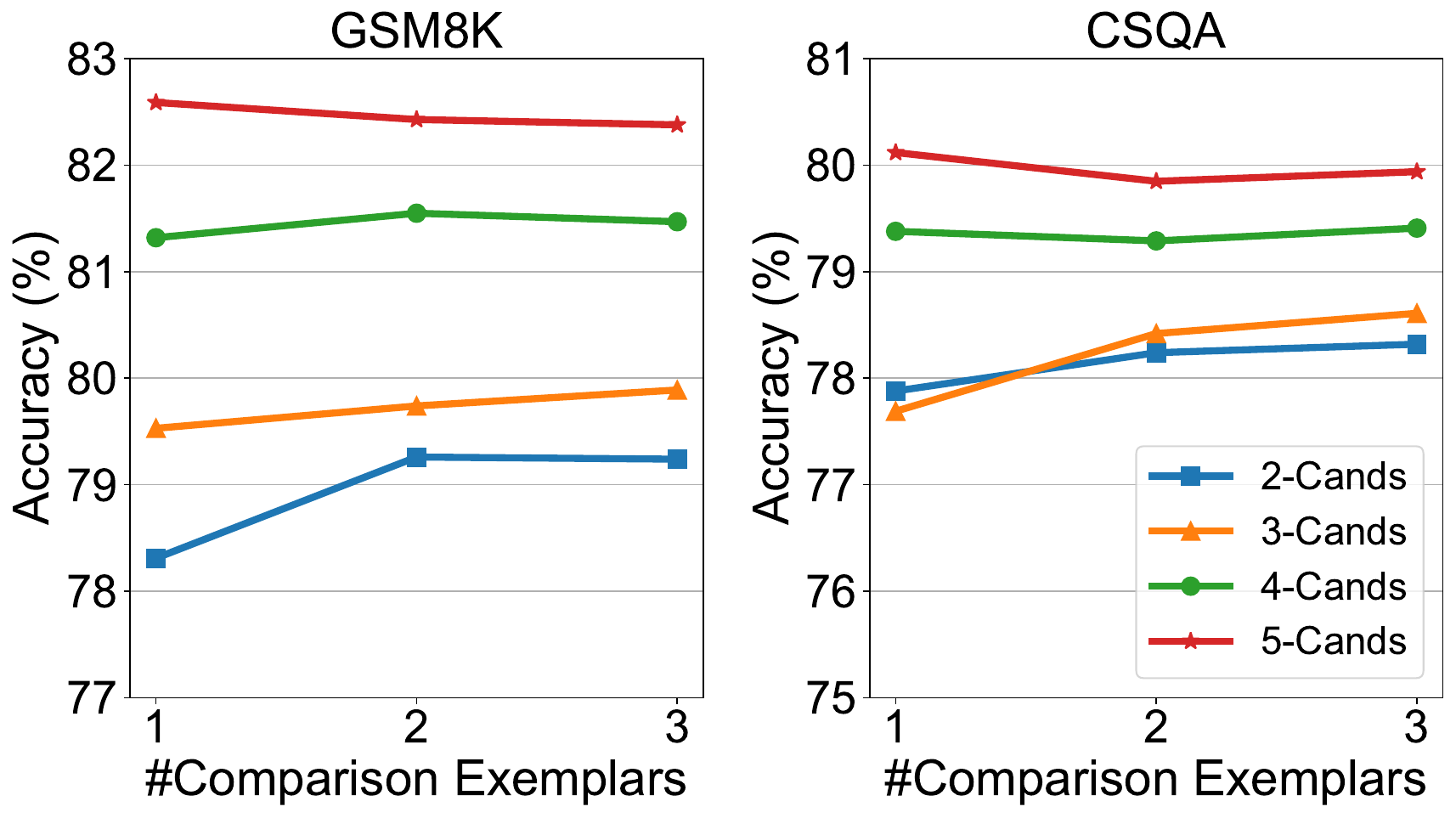}
    \caption{Test accuracy with varying complexity and numbers of comparison exemplars. The results are obtained on GSM8K (left) and CSQA (right) using \texttt{gpt-3.5-turbo-16k}.}
    \label{fig:complexity}
\end{figure}

\paragraph{Exemplar complexity is much more important than quantity.} Beyond exemplar correctness, we delve into the influences of complexity and quantity on ranking performance. Intuitively, ranking an expansive and diverse set of candidates inherently possesses greater complexity. This complexity may serve as a reflection of the depth and detail involved in the ranking process. We utilize the count of unique candidates involved in a single comparison exemplar as an indicator of its complexity. We perform ranking over 5 candidates using \texttt{gpt-3.5-turbo-16k}, which supports up to 16K tokens. For instance, Figure \ref{fig:complexity} presents the results from the GSM8K test set. "N-Cands" denotes an exemplar that illustrates the ranking process across $N$ different candidates. The results reveal that the complexity of exemplars is much more important than the quantity. Remarkably, we find that employing a single \textit{complex} exemplar is more effective than using multiple \textit{simple} exemplars.

\subsection{Impact of Candidate Answers}
We have demonstrated that RankPrompt is robust to the inconsistency in candidate answers in Section \ref{sec:inconsistency_exp}. Here, we further investigate the behaviors of different methods by varying the number and order of candidates.

\begin{figure}[t]
    \centering
    \includegraphics[scale=0.25]{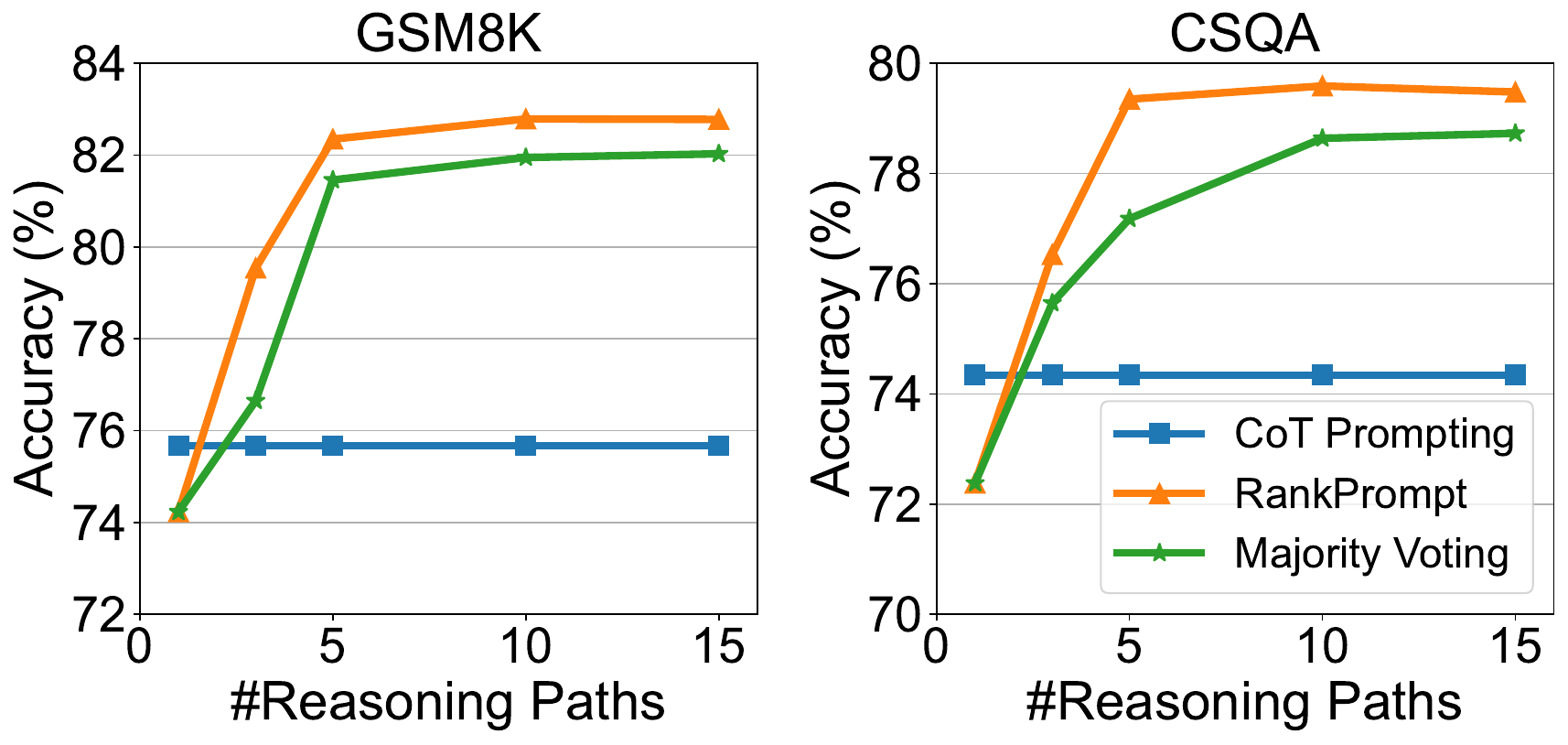}
    \caption{Test accuracy measured against varying numbers of reasoning paths. The results are obtained on GSM8K (left) and CSQA (right) using \texttt{gpt-3.5-turbo-16k}. CoT Prompting uses greedy decoding, while others employ sampling (temp=0.7).}
    \label{fig:candidates}
\end{figure}

\paragraph{Using more candidates offers minor benefits.}
\label{exp:num_candidates}
In our main experiments, we opt for 5 candidates, partially due to the input length constraint of LLMs. For instance, \texttt{gpt-3.5-turbo} has a 4096-token limit. Here, we explore the impact of increasing the number of candidates using \texttt{gpt-3.5-turbo-16k}. We evaluate CoT Prompting, Majority Voting, and RankPrompt on the test sets of GSM8K and CSQA, varying the number of sampled reasoning paths (1, 3, 5, 10, 15). As plotted in Figure \ref{fig:candidates}, both RankPrompt and Majority Voting show improved performance with more candidates, but the gains plateau beyond 5 reasoning paths. While further increasing the number of candidates offers slight improvements, it also significantly raises the cost. Hence, we recommend using 5 candidates to make trade-offs between performance and cost.

\begin{figure}[t]
    \centering
    \includegraphics[scale=0.25]{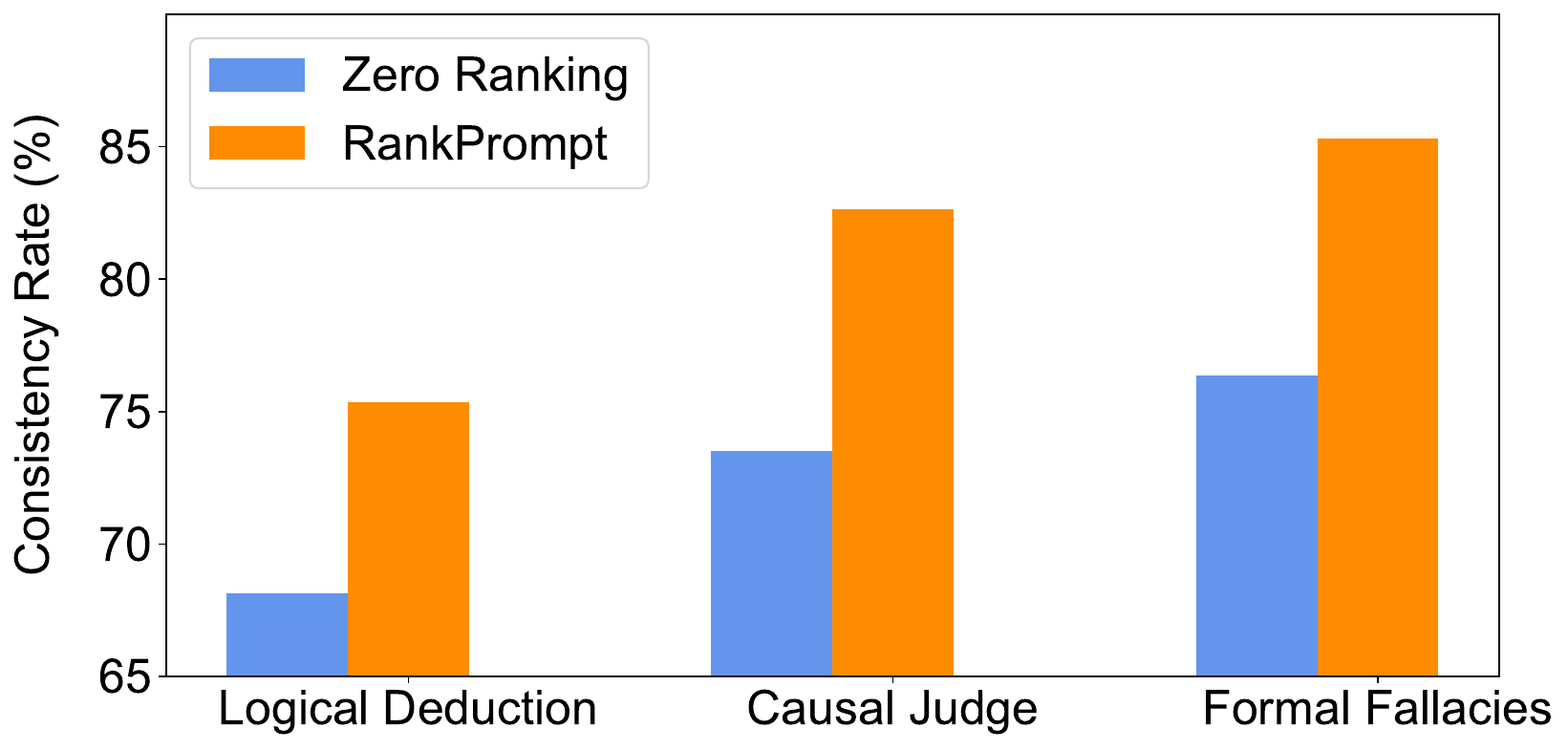}
    \caption{Consistency rates of Zero Ranking and RankPrompt when ranking 5 candidates shuffled 3 times. The results are obtained with \texttt{gpt-4}.}
    \label{fig:robustness}
\end{figure}

\paragraph{RankPrompt is robust to the ordering of candidates.}
A good evaluator should exhibit robustness against variations in the order of candidate answers. In this section, we investigate the robustness of different ranking methods on the challenging BBH tasks. We employ the identical experimental setup specified in Section \ref{sec:bbh_exp} and run the ranking process 3 times, with candidate orderings being shuffled each time. Instead of reporting the overall accuracy, which would gain from increasing individual reasoning paths, we focus on the prediction consistency across different methods. Specifically, we regard a ranking as \textit{consistent} if it remains unchanged across all 3 iterations. As depicted in Figure \ref{fig:robustness}, RankPrompt exhibits greater robustness compared to Zero Ranking when confronted with variations in candidate orders. Specifically, RankPrompt produces consistent rankings ranging from 75\% to 85\% of the time. These results demonstrate that RankPrompt is a reliable and robust judge for complex reasoning tasks.

\input{tables/error_types}

\subsection{Error Analysis}
To gain further insights into how RankPrompt enhances the reasoning performance of language models, we manually analyze the errors made by RankPrompt and CoT Prompting on AQUA-RAT. We utilize the same error categorizations as in \cite{sawada2023arb} for the qualitative analysis of the results in \ref{table:main_results}. In total, RankPrompt produces 72 errors, while CoT Prompting accumulates 105 errors. We find that RankPrompt mitigates all types of errors identified in CoT Prompting. Interestingly, both CoT Prompting and RankPrompt make a few calculation errors (15 vs. 9). RankPrompt significantly reduces errors caused by wrong approaches (from 42 to 27) but proves less effective in mitigating the impact of misinterpretation (from 17 to 14).

\section{Conclusion}
We have presented RankPrompt, a novel prompting method for selecting the optimal output from a diverse set of reasoning paths generated by LLMs. This method systematically steers LLMs to compare potential answers, leveraging step-aware comparison instructions and automated exemplars. This approach confers three primary advantages: (1) it eliminates the need for additional models and human annotations, (2) it achieves strong performance across a broad spectrum of reasoning and automatic evaluation tasks, and (3) it is robust to inconsistent reasoning paths. Our comprehensive evaluation underscores that the precision and complexity of comparison exemplars play a critical role in ranking performance. Collectively, our findings position RankPrompt as an effective strategy to enhance the reasoning capabilities of LLMs.

\section*{Acknowledgement}
This work was supported in part by the National Science Foundation of China (No.62276056), the Natural Science Foundation of Liaoning Province of China (2022-KF-16-01), the Fundamental Research Funds for the Central Universities (Nos. N2216016 and N2316002), the Yunnan Fundamental Research Projects (No. 202401BC070021), and the Program of Introducing Talents of Discipline to Universities, Plan 111 (No.B16009). The authors would like to thank anonymous reviewers for their insightful comments.

\section*{Limitations}
Despite the impressive performance of our method, its experiments has been limited to proprietary language models. The lack of publicly accessible training details for these models creates a significant barrier for researchers interested in pursuing enhancements from a modeling standpoint. In the future, we will enhance the ranking capabilities of open-source models like LLaMA \cite{Touvron2023Llama2O, Touvron2023LLaMAOA} and Falcon \cite{penedo2023refinedweb}. Learning from the explanations behind GPT-4's ranking decisions offers a promising path for exploration. Additionally, while comparison exemplars in prompts improves performance, they also significantly increases the context size, leading to more expensive API calls. A potential solution is to condense the candidate paths by summarizing their key points.

\nocite{*}
\section{Bibliographical References}\label{sec:reference}

\bibliographystyle{lrec-coling2024-natbib}
\bibliography{lrec-coling2024-example}

\bibliographystylelanguageresource{lrec-coling2024-natbib}
\bibliographylanguageresource{languageresource}

\end{document}

%% file: tables/intro.tex
\begin{table}[ht]
\small
\centering
\footnotesize
% \vspace{8pt}
\begin{tabular}
{p{0.46\textwidth}}
\toprule
{\textbf{Question:} Bob spends \$27000 distributed between 3 schools to buy books. He can buy 100 books for \$500. How many books can he buy per school?} \\
\midrule
{\textbf{Answer 1:} Bob spends a total of \$27000 on books. Each book costs \$500. \textcolor{red}{Therefore, he can buy \$27000 / \$500 = 54 books in total.} Since he is distributing the books between 3 schools, he can buy 54 / 3 = 18 books per school.   \textcolor{red}{\ding{56}}} \\
\midrule
{\textbf{Answer 2:} Bob can buy 27000 / 500 = 54 sets of 100 books.
So, Bob can buy 54 x 100 = 5400 books.
Therefore, he can buy 5400 / 3 = 1800 books per school. Answer: 1800.  
 \textcolor{green}{\ding{52}}}  \\
\midrule
{\textbf{Answer 3:} Bob can buy 100 books for \$500, so the cost of one book is \$500 / 100 = \$5.
Bob spends \$27000, so he can buy 27000 / 5 = 5400 books.
\textcolor{red}{Therefore, Bob can buy 5400 books per school.}   \textcolor{red}{\ding{56}}} \\
\bottomrule
\end{tabular}
\caption{An example from GSM8K \cite{Cobbe2021TrainingVT}. Answer 2 is correct, while others make invalid inferences or miss steps in their reasoning process (marked in red). In this case, there is no major answer among all candidates.}
  \label{tab:intro}
\end{table}

%% file: tables/ranking_template.tex
\begin{table}[t]
\small
\begin{tcolorbox}

[Comparison Instruction]

You are provided with a question and a series of potential responses. 

Your assignment involves a systematic, step-by-step comparison of the reasoning paths embedded within each response. 

This entails a thorough evaluation of each step's correctness and logical consistency. 

After completing this all-encompassing assessment, rank the responses in accordance with the soundness of their respective reasoning paths. 

Finally, select the best response and \textcolor[rgb]{0.85,0,0}{present it on a separate line as the optimal solution}.

[Comparison Example]

\textcolor[rgb]{0,0,0.85}{\{Comparison Exemplars\}}

[Question]

\textcolor[rgb]{0,0,0.85}{\{Question\}}

[Candidate Answers]

\textcolor[rgb]{0,0,0.85}{\{Candidates\}}

[Comparison]

Let's compare the answers step by step.

\end{tcolorbox}
\caption{The ranking template of RankPrompt. It instructs LLMs to compare candidate answers step by step and output in a specific format (marked in red).}
\label{tab:ranking_template}
\end{table}

%% file: tables/main_results.tex
\begin{table*}[t]\centering
% \vspace{-3mm}
\setlength{\tabcolsep}{3.7pt}
% \vspace{2.8mm}
\begin{tabular}{lccccccccc}\toprule
\multicolumn{1}{l}{\multirow{2}*{{\textbf{Method}}}}  &\multicolumn{4}{c}{{\textbf{Arithmetic}}}  &\multicolumn{3}{c}{{\textbf{Commonsense}}} &\multicolumn{1}{c}{{\textbf{Symbolic}}} &\multirow{2}*{{\textbf{Avg.}}}\\
\cmidrule(r){2-5}
\cmidrule(r){6-8}%
\cmidrule(r){9-9}%
\multicolumn{1}{c}{~} & AQUA & GSM8K & SVAMP & ASDiv & StrategyQA & CSQA & ARC & LastLetter \\

\midrule
CoT Prompting  & 58.51 & 75.89 & 80.10 & 87.07 & 72.88 & 74.12 & 85.26 & 74.40 & 76.03  \\
\midrule
Majority Voting & 62.60 & 81.27 & 83.80 & 88.36 & 74.06 & 77.48 & 87.31 & 76.40 & 78.91  \\
Direct Scoring   & 63.39 & 80.14 & 82.60 & 88.69 & 73.14 & 78.36 & 87.20 & 76.60 & 78.77  \\
Zero Ranking & 67.72 & 79.98 & 83.30 & 89.36 & 74.55 & 78.21 & \textbf{87.57} & 74.80 & 79.44  \\
RankPrompt & \textbf{71.65} & \textbf{82.43} & \textbf{84.30} & \textbf{90.12} & \textbf{76.07} & \textbf{79.20} & 87.42 & \textbf{77.60} & \textbf{81.10}  \\
\midrule
Oracle & 79.53 & 91.05 & 91.40 & 94.18 & 85.37 & 85.26 & 92.83 & 86.40 & 88.25  \\

\bottomrule
\end{tabular}

\caption{Comparisons of the accuracy on 8 reasoning tasks with \texttt{gpt-3.5-turbo}. CoT Prompting uses greedy decoding (temp=0), while other methods sample 5 candidates (temp=0.7). The best performance for each task under the same settings is shown in \textbf{bold}.}
\label{table:main_results}
\end{table*}

%% file: tables/bbh_results.tex
\begin{table}[t!]
\centering
\footnotesize
\begin{tabular}{lccc}
\toprule
\multicolumn{1}{l}{{\textbf{Method}}} & 
\multicolumn{1}{c}{{\begin{tabular}[c]{@{}c@{}}{\textbf{Logical}}\\ {\textbf{Deduction}}\end{tabular}}} & \multicolumn{1}{c}{{\begin{tabular}[c]{@{}c@{}}{\textbf{Causal}}\\ {\textbf{Judge}}\end{tabular}}} & \multicolumn{1}{c}{{\begin{tabular}[c]{@{}c@{}}{\textbf{Formal}}\\ {\textbf{Fallacies}}\end{tabular}}} \\
\midrule
CoT Prompting & 57.60 & 69.52 & 76.80 \\
\midrule
Majority Voting & 62.40& 72.19  & 82.40 \\
Direct Scoring & 61.20& 71.12  & 81.60 \\
Zero Ranking & 63.70 &72.51 &  83.20 \\
RankPrompt & \textbf{66.80}& \textbf{74.73}  & \textbf{84.40} \\
\midrule
Oracle & 90.00 & 79.14 & 92.40  \\
\bottomrule
\end{tabular}
\caption{Test accuracy on 3 challenging BBH tasks using \texttt{gpt-4} over 5 candidates.}
  \label{tab:bbh_results}
\end{table}

%% file: tables/alpaca_results.tex
\begin{table}[t]
\centering
\footnotesize
\begin{tabular}{lcr}
\toprule
\multicolumn{1}{l}{{\textbf{Method}}} & 
% \multicolumn{1}{c}{{\begin{tabular}[c]{@{}c@{}}{\textbf{Human}}\\ {\textbf{Agreement}}\end{tabular}}} 
\textbf{Human Agreement}
& \multicolumn{1}{r}{{\textbf{Price}}} \\
\midrule
Inter-Human & 65.70 & \$241.50 \\
\midrule
Direct Scoring & 64.48 & \textbf{\$11.19} \\
AlpacaFarm & 67.22& \$12.35 \\
Alpaca Evaluator & 70.13& \$14.23 \\
Zero Ranking & {71.67}& \$16.74 \\
RankPrompt & \textbf{74.33} & \$19.18 \\
\bottomrule
\end{tabular}
\caption{Human agreements and cost on the test set of AlpacaEval using \texttt{gpt-4}. Inter-Human denotes the average results of human annotators.}
  \label{tab:alpaca_results}
\end{table}

%% file: tables/error_types.tex
\begin{table}[t]
\centering
\setlength{\tabcolsep}{3.3pt}
\footnotesize
\begin{tabular}{lcc}
\toprule
\multicolumn{1}{l}{{\textbf{Error Type}}} & 
\textbf{CoT Prompting} & 
\textbf{RankPrompt}  \\
\midrule
Calculation Error & 15 & 9  \\
Wrong Approach & 42 & 27  \\
Misinterpretation & 17 & 14  \\
Logical Error & 31 & 22 \\
\midrule
Total Errors & 105 & 72 \\
\bottomrule
\end{tabular}
\caption{Error statistics on the AQUA-RAT dataset using \texttt{gpt-3.5-turbo}.}
  \label{tab:error_types}
\end{table}